# A Review on Automatic License Plate Recognition System

Satadal Saha
Dept. of Electronics and Communication Engg.
MCKV Institute of Engineering
Liluah, Howrah

*Abstract-* **Automatic License Plate Recognition (ALPR) is a challenging problem to the research community due to its potential applicability in the diverse geographical condition over the globe with varying license plate parameters. Any ALPR system includes three main modules, viz. localization of the license plate, segmentation of the characters therein and recognition of the segmented characters. In real life applications where the images are captured over days and nights in an outdoor environment with varying lighting and weather conditions, varying pollution level and wind turbulences, localization, segmentation and recognition become challenging tasks. The tasks become more complex if the license plate is not in conformity with the standards laid by corresponding Motor Vehicles Department in terms of various features, e.g. area and aspect ratio of the license plate, background color, foreground color, shape, number of lines, font face/ size of characters, spacing between characters etc. Besides, license plates are often dirty or broken or having scratches or bent or tilted at its position. All these add to the challenges in developing an effective ALPR system.**

*Integrated Traffic Management System* (ITMS) has been installed in most of the developed countries with an objective to track on-road traffic violations, using surveillance cameras and intelligent image analytic softwares. The different components of typical ITMS involve 1) monitoring of vehicle speed on road, 2) automatic estimation of traffic volumes at different traffic intersections, 3) synchronized signalling system in city roads, 4) detection/ localization of illegal parking and detection of wrong way traffic, 5) detection of stop-line violating vehicles etc. In recent years, it has now become a growing need by different traffic monitoring authorities in India for automatic identification of vehicles that has violated traffic signal at a road crossing. The purpose of any such Stop Line Violation Detection System (SLVDS) is to track down the vehicles that have violated the traffic signals at a road crossing. The systems may also be implemented at toll plaza, car parking areas and in security zones for automatic recognition of license number of the vehicles that have entered into the area for specific purposes. A real life, ALPR system needs to address various outdoor environmental factors discussed below.

*Camera related issues*: Mounting of the camera is also difficult in outdoor environment. Once mounted, the front face of the camera is usually angled towards the front face of the road. This often introduces some skew in the captured images. Dust pollution, vibration of the camera, wind turbulence and heavy vehicular movement on the road make poor quality images.

*Environment related issues*: When the ALPR system is installed on road, the acquired images change due to various factors. For example, over the day and night, due to the changes in ambient lighting condition, the overall brightness and contrast of the images change significantly. The ALPR system should also run over the different seasons, like summer and monsoon. In rainy (monsoon) season the roads get wet thereby changing the background. This change in image quality due to seasonal variation should also be addressed in case of an outdoor ALPR system.

*Traffic related issues*: The performance of an APLR system mostly depends on the vehicular traffic density and the population density of the place where the system is to be installed. High vehicular traffic density may sometimes cause non-aligned vehicles at some crossing as well as partial or full occlusion of the license plates by other vehicles.

*License plate related issues*: There are standard specifications of the license plate in every country. The owner of the vehicle should adhere strictly to those specifications to obey the traffic rules and regulations. Lack of standardization in some country (or in specific states of the country) or reluctant nature of administration in enforcing the standard specifications of the license plate makes the task of automatic localization of the license plate very difficult. In India, people often remain indifferent to adhering to the standard values of the aforementioned regulations. This makes the problem of automatic identification and recognition of the license plate more difficult.

The dataset for the current work has been developed for a project on SLVDS [1] as a part of ITMS implemented by the traffic monitoring authority of a major metro city in India. Multiple surveillance cameras were installed to detect the vehicles violating the stop-line at a traffic intersection during red traffic signal. During the data collection phase, the system was run through several days/nights in unconstrained outdoor environments with varying lighting and weather conditions, high pollution levels and wind turbulences. It was made to run throughout the two seasons: summer and monsoon. More than 30,000 video snapshots were taken at a rate of 25 fps and stored as 24-bit color bitmaps and with a resolution of $704 \times 576$ pixels. A total of 4717 frontal vehicular snapshots are selected from 13 different camera view points for preparing the dataset.

*Binarization* plays a key role in license plate localization, segmentation and recognition. It is the process of converting a grey scale image (popularly known as *multi-tone* image) into a black-and-white image (popularly known as *two-tone* image). There are several binarization techniques [2–4] most of which deal with document images with text and graphics and fail miserably to address the complexity of the current problem. A review on image binarization is reported by Nandy et *al.* [5]. Saha et *al.* proposed two histogram equalization based methods [6–8] that assign membership values to each pixel depending on their intensity value.

*License plate localization* is the process of identifying the region within the scene image where the license plate the vehicle actually lies. Several methods have been proposed in different literatures regarding license plate localization. One method [9] on color based segmentation was an initial step towards the localization of the license plate for commercial vehicles providing an *f-measure* value of 84.66%. Spanish license plate is localized and recognized in [10] using connected component labeling algorithm of different binarized images generated from different thresholds. Kwasnicka et *al.* [11] proposed a hybrid method to localize the license plate. During the localization phase, the position of the characters is used in [12]. It assumes that no significant edge lies near the license plate boundary and the inside characters are disjoint. Another method [13] based on color feature based neural network has been proposed for localization of license plate of all types of vehicles. It performs reasonably good compared to the previous method, providing an *f-measure* value of 91.66%. A work on localization of Iranian license plate is done in [14], [15] using mean shift algorithm for localization of license plate giving satisfactory result for license plates having color different from the body color. In another work [16], [17], the vertical edge attributes of English characters has been used as features. A vertical edge based multi-stage approach has been proposed for license plate localization. The method provides a reasonably high $f-measure$ of 94.29%. A Hough transform based license plate localization method [18] has been developed which provides an *f-measure* of 92.18%. A hybrid method [19], [20] consisting of vertical edge based method and Hough transform based method has been proposed by Saha et *al.* generating an average *f-measure* of 94.79%. Another hybrid method [21] consisting of iterative edge map and Hough transform has been proposed by Saha et *al.* giving an *f-measure* of 95.83%.

*Segmentation* plays a vital role for recognition of character through an Optical Character Recognition system. A good segmentation technique results in a better recognition compared to a bad segmentation technique with the same recognizer. The traditional segmentation techniques primarily aimed for document image segmentation do not perform well when applied on camera captured images in an unconstrained outdoor environment. In [22], a license plate localization and character segmentation method is proposed.

Hegt et *al.* [23] proposed a projection based method for license plate character segmentation. Haris et *al.* [24] proposed a hybrid multidimensional image segmentation algorithm which combines edge and region-based techniques through the morphological algorithm of watersheds. A technique based on hierarchical segmentation of touching characters from camera captured images is based on the method of binarization at the different levels of object hierarchy. It efficiently segments very loosely connected neighboring characters with an accuracy of 92.22%. It could not segment very strongly connected printed characters. To solve these types of cases, another method has been developed for segmentation of touching characters using a fuzzy water flow based method. The method has become useful by segmenting unintentional but strongly touching neighboring characters with an accuracy of 97.23%.

*Recognition* is the final module of an ALPR system. Different objects or patterns can be classified by means of finding discriminating feature set exhibiting different properties of the objects or patterns. More specifically, features are the quantitative measures of a pattern, that as a whole impose a membership value of the pattern to a particular class. The purpose of feature extraction is to reduce the amount of data required for representation of the properties of different patterns. The feature set should be efficient in discriminating closely separated properties of patterns and it should not contain redundant features within it. A two stage hybrid recognition system combining statistical and structural features is proposed in [25]. Llorens et *al.* [10] used HMM to recognize the characters. For recognition of the license plate characters, multi-layer perceptron (MLP) is used as a classifier [19]. Quad Tree based Longest Run (QTLR) feature set with Center of Gravity (CG) based partitioning is used to train the network. Overall plate level accuracy with no recognition error is found to be 92.75%. The character level accuracy of 98.76% has been achieved in the proposed algorithm. Some post-processing techniques are normally been employed following the syntax of the license number. Saha et *al.* [26] reported a complete license plate recognition system in Indian scenario.